\documentclass[letterpaper]{article} %
\usepackage{aaai23}  %
\usepackage{times}  %
\usepackage{helvet}  %
\usepackage{courier}  %
\usepackage[hyphens]{url}  %
\usepackage{graphicx} %
\usepackage{natbib}  %
\usepackage{caption} %
\usepackage[utf8]{inputenc} %
\usepackage[T1]{fontenc}    %
\usepackage{url}            %
\usepackage{booktabs}       %
\usepackage{amsfonts}       %
\usepackage{nicefrac}       %
\usepackage{microtype}      %
\usepackage{xcolor}         %

\usepackage{amsfonts}       %
\usepackage{nicefrac}       %
\usepackage{microtype}      %
\usepackage{xcolor}         %

\usepackage{graphicx}
\usepackage{subfigure}
\usepackage{amsmath}
\usepackage{bm}
\usepackage{array,multirow}
\usepackage{float}

\usepackage{newfloat}
\usepackage{listings}
\DeclareCaptionStyle{ruled}{labelfont=normalfont,labelsep=colon,strut=off} %
\floatstyle{ruled}
\newfloat{listing}{tb}{lst}{}
\floatname{listing}{Listing}
\title{A Data Source for Reasoning Embodied Agents}
\author {
    Jack Lanchantin,
    Sainbayar Sukhbaatar,
    Gabriel Synnaeve,\\
    Yuxuan Sun,
    Kavya Srinet,
    Arthur Szlam
}
\affiliations {
    Meta AI\\
    \{jacklanchantin, sainbar, gab, yuxuans, ksrinet, aszlam\}@meta.com
}

\begin{document}

\maketitle

 \begin{abstract}
Recent progress in using machine learning models for reasoning tasks has been driven by novel model architectures, large-scale pre-training protocols, and dedicated reasoning datasets for fine-tuning.  In this work, to further pursue these advances, we introduce a new data generator for machine reasoning that integrates with an embodied agent.  The generated data consists of templated text queries and answers, matched with world-states encoded into a database.  The world-states are a result of both world dynamics and the actions of the agent.  We show the results of several baseline models on instantiations of train sets.  These include pre-trained language models fine-tuned on a text-formatted representation of the database, and graph-structured Transformers operating on a knowledge-graph representation of the database.  We find that these models can answer some questions about the world-state, but struggle with others. These results hint at new research directions in designing neural reasoning models and database representations. 
Code to generate the data will be released at \url{github.com/facebookresearch/neuralmemory}.
\end{abstract}
 \section{Introduction}

Advances in machine learning (ML) architectures \cite{vaswani2017attention}, large datasets and model scaling for pre-training \cite{radford2019language,chowdhery2022palm,zhang2022opt}, modeling approaches \cite{nye2021show, wang2022self}, and multiple dedicated reasoning datasets \cite{yang2018hotpotqa, Hudson2019GQAAN, petroni2020kilt, vedantam2021curi} have driven progress in both building models that can succeed in aspects of ``reasoning'' and in automatically evaluating such capabilities. This has been evident particularly in the text setting, but also in computer vision \cite{krishnavisualgenome, Johnson2017CLEVRAD, Yi2020CLEVRERCE}.

In parallel, the last decade has seen advances in the ability to train embodied agents to perform tasks and affect change in their environments.  These have been also powered in part by data, with many environments made available for exploring modeling approaches and benchmarking.   In particular, with respect to ``reasoning'' in embodied agents, there have been works showing that adding inductive biases to support reasoning can lead to improved performance with end-to-end training \cite{zambaldi2018deep} and other works have shown how models can be augmented with extra supervision \cite{zhong2019rtfm}.

Recently, several works have shown how large language-model pre-training can be used to affect planners for embodied agents \cite{huang2022language, ahn2022can}. More generally, symbolic representations can be a hub for connecting perception, memory, and reasoning in embodied agents. 

\begin{figure}[t!]
\centering
\includegraphics[width=0.9\columnwidth]{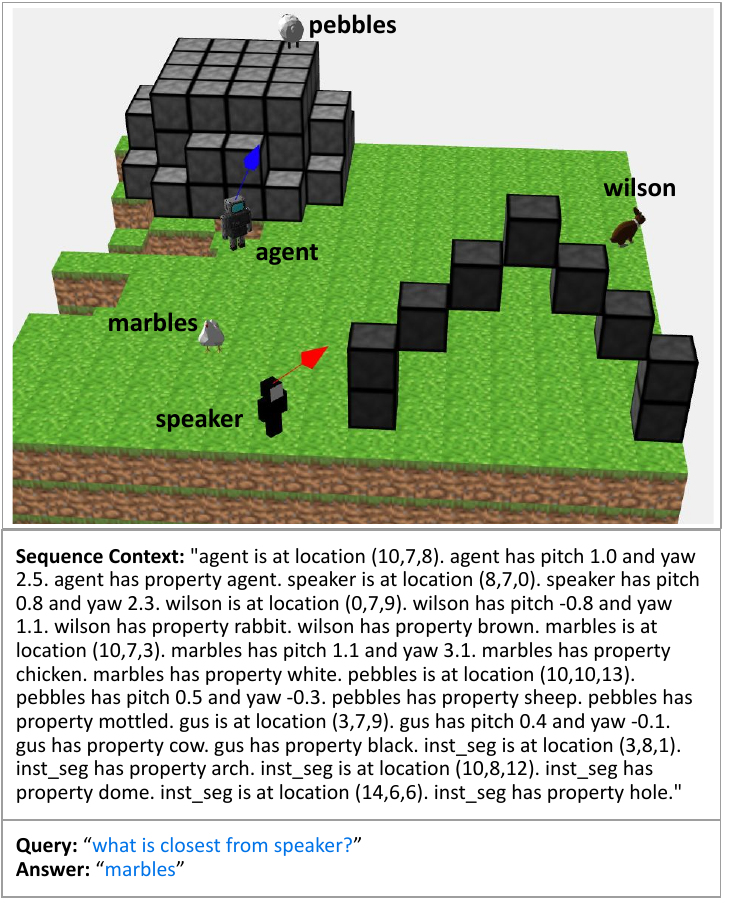}
\caption{ 
(Top): Example of a generated scene in our 3d gridworld. %
(Middle): Given the 3d scene, we can convert the information in the render to a text or structured representation. Here we show the text Sequence Context representation. ``inst\_segs'' represent the block items such as structures or holes.
(Bottom): For a particular scene, we can generate a wide variety of queries. Here we show and example of a distance query asking which object is the closest to the speaker, which is the chicken named ``marbles''.
}
\label{fig:data_sample} 
\end{figure}

However, the growing literature in NLP reasoning models is missing data grounded in a dynamic and agent-alterable world. Models trained on traditional text datasets struggle to handle physically grounded queries such as those that involve geometric reasoning. 
In other words, recent large language models trained on internet data are not well equipped to simple questions about a physical environment such as  ``who is to my left?''. 
Grounding large-language models may allow them more powerful reasoning; and vice versa, may help us use them as agent controllers.

In this work we describe a data source (i.e., a toolbox to generate data) designed to help train ML models grounded in a physical environment, allowing them to make the connection between perception, memory, and reasoning.  It consists of context-question-answer triples, where the context corresponds to a dynamic and agent-affected 3d gridworld, and the questions may involve temporal or spatial reasoning, as well as questions about the agent's own actions. 
A sample generated from our data source is shown in Fig.~\ref{fig:data_sample}.

While the environment allows rendering the world-context as a sequence of images, 
one of our goals is to support research toward answering the question ``what are good formats for agent memory systems?''.  In pursuit of this, we abstract the context to a database format that does not require any perceptual modules, and provide code for converting the database into a templated text dump, as demonstrated in Fig.~\ref{fig:data_sample} (right top). Here, the order of the facts within each timestep are written randomly,  and sequentially written according to the timesteps. Our hope is that the data source can be used for augmenting the training (or allowing the assembly) of reasoning embodied agents by bringing to bear the advances in reasoning in language models, or as a supplement to training language models with grounding from embodied agents.

We train baseline neural models to represent the database and process the queries. These include finetuning pre-trained language models on the text version of the database, and Transformers that input the structured database directly.  We find that while certain queries are easily solved by these baselines, others, in particular - those having to deal with spatial geometry, are more difficult.

In short, the contributions of this paper are:
\newline
{\bf Environment}: We introduce an environment for embodied agents and a data source for generating data to train agents in this environment (detailed in the Environment, Queries, and Data section). We provide the code to generate world contexts, as well as complex queries. We hope this will aid researchers to isolate and tackle difficult problems for reasoning for embodied agents. 
\newline{\bf Baselines}: We evaluate the abilities of baseline models to answer queries in this environment 
(Experiments section). 
We compare different representations of the world context, including a pure text based representation as well as a more structured representation.

\section{Environment, Queries, and Data} \label{data}
We propose a context-question-answer data generator for embodied agents.  In this section, we outline the context, or environment we generate data in, the types of queries we create for the agent to solve, and specifics of the data samples.

\subsection{Environment} We work in a finite three-dimensional gridworld.  There is a primary agent, zero or more other agents, zero or more get/fetchable items, zero or more placeable/breakable blocks. The other agents come in two types: they might represent human ``players'' that can give commands to the agent and have the same capabilities as the agent, or animate non-player characters (NPCs) that follow simple random movement patterns.  The placeable/breakable blocks have colors and integer grid coordinates; all other objects have float coordinates. Animate objects (the players, NPCs, and the agent) have a yaw and pitch pose representing the location they are looking, in addition to three location coordinates.  

To build a scene, we generate some random objects (spheres, cubes, etc.),  randomly place a number of NPCs, a player, and an agent.  With some probability, the agent executes a command (to either: build an object, destroy an object, move to a location, dig a hole, follow an NPC, etc.)  The execution of a command is scripted; the task executor is from the Minecraft agent in \cite{pratik2021droidlet}.  
Whether or not the agent executes a task, the world steps a fixed number of times (so, e.g., NPCs may move or act).  In the experiments described below, the world is fully observed at a fixed number of temporal snapshots, and all poses, object locations and NPC movements are recorded. However, not every world step is snapshotted, so the total sequence is not fully observed.

Following \cite{pratik2021droidlet}, the environment is presented to the agent as an object-centered key-value store.  Each object, NPC, and the agent's self have a ``memid'' keying a data structure that depends on the object type, and may contain string data (for example a name) or float or integer data (e.g. the pose of an NPC).  The key-value store also has subject-predicate-object triples (e.g. "memid has tag mottled"); the triples themselves also have a unique memid as key.

\begin{figure*}
\centering
\includegraphics[width=0.83\textwidth]{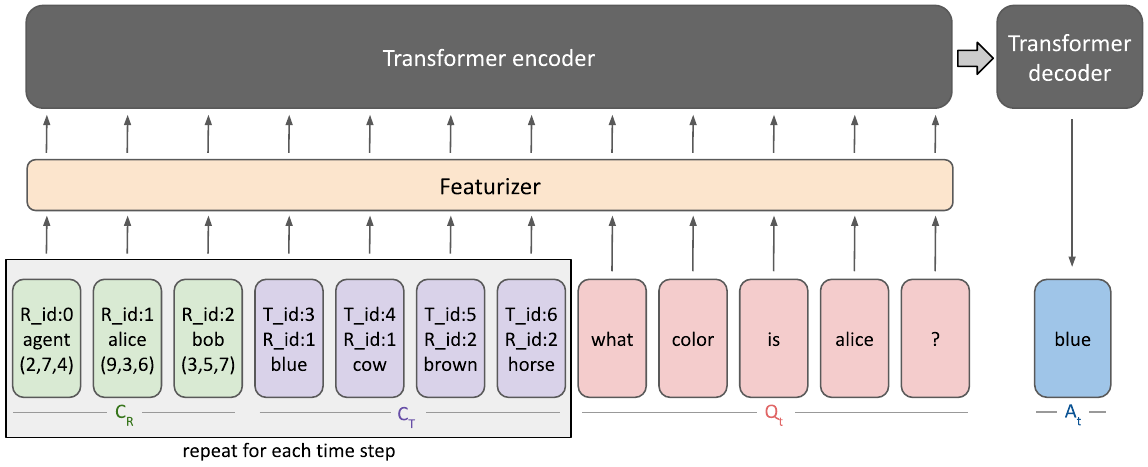}
\caption{ 
Structured context + Transformer model. The bottom left demonstrates the structured representation of the 3d gridworld, where R\_id is the unique reference object identifier, and T\_id is the unique triple property identifier (which connects to one of the reference objects via R\_id). Context nodes ($C_{\{R,t\}}$) and query tokens ($Q_{t}$) are first featurized with learnable embedding layers. We process the featurized context and query jointly with a Transformer encoder that considers the context structure via relational embeddings 
($r_{ij}$). 
Finally, a text decoder predicts tokens, and a memid decoder predicts relevant context memids (not pictured).}
\vspace{-3pt}
\label{fig:structured_model}
\end{figure*}

The generated scene presented as the key-value store defines the agent's context, $C$. In our experiments below, we represent this context in one of two ways. The first is a text sequence ($C_{t}$), where for each snapshot step, all objects and their properties in the key-value store are flattened into a templated language, as shown in Fig.~\ref{fig:data_sample} (right). Multiple time snapshots of the context are represented by appending each successive event in the sequence.   While this is in some sense simplistic, it allows ready application of large pre-trained language models for ingesting the context; and this kind of approach has been shown to be effective in \cite{thorne2021database, liu2021tapex}. 

Alternatively, we can leverage the relational properties of the context by representing it as a graph where objects and properties are nodes, and the connections between them are the edges. For example, if we know that ``bob'' is a horse and is the color brown, the node for bob connects to a ``horse'' and a ``brown'' object. Specifically, this ``Sequence Context'' contains reference object nodes ($C_{R}$), which are the object instantiations, and triple nodes ($C_{T}$), which are the properties of the reference objects. Each reference object node holds the following information:
\textit{reference\_object\_hash} (R\_id) is a unique identifier for each reference object,
\textit{reference\_objects\_words} holds identifier words of the object such as its name, and 
\textit{reference\_objects\_float} is the floating point properties of the object such as its (x, y, z) coordinates and pitch/yaw. These are combined into a single node 
(detailed in the Models section).
Similarly, each property triple is composed of the following elements: 
\textit{triples\_hash} (T\_id) contains a unique identifier for the triple, as well as the reference object hash that it is linked to, and
\textit{triples\_words} are a descriptive text of a reference object's property such as ``has\_color blue''.
These are also combined into a single node. We therefore have nodes of both reference objects and triples, and the hashes encompass the edges or relationships between them.

We do not consider the text sequence or graph-structured representations to be canonical.  On the contrary, our goal is stimulate research into what the correct representation of world state should be to allow easier model training and transfer between agents. We simply provide these two representations as examples and baselines.

Given the determined set of snapshots, queries are designed to be answerable. That is, all information is known to answer the questions.   We leave to future work making ambiguous queries, but note here that it is not difficult to purposefully build and record queries that could theoretically be answered in some scene but cannot be answered in the particular scene instantiation, for example because they refer to an event that occurred between snapshots.  It would otherwise be easy to restrict full observability within each snapshot.
\subsection{Queries}
\begin{table*}[t!]
\fontsize{9}{9}
\centering
\resizebox{0.87\linewidth}{!}{
\centering
\begin{tabular}{lll}
\toprule
Query Class & Clause types & Example     \\ \midrule
Property      & name& what are the properties of the objects that have the name alice?  \\
   & tag & what are the names of the objects that has the property brown?    \\
   & absolute cardinal     & what are the locations of the objects where the x coordinate is less than 4?        \\ \midrule
Temporal    & cardinal     & what is the name of the object that increased x the most?  \\
   & relative     & what is the name of the object that moved to my left the most?    \\
   & \textit{farthest moved object} & what is the name of the object that moved the farthest?  \\
   & \textit{location at time}      & what was the location of bob at the beginning?  \\
   & \textit{action}       & what did you do?     \\ 
   & \textit{object tracking}       & where would the ball be if i moved to (4,7,2)?     \\ 
   \midrule
Geometric   & absolute distance     & what is the count of the objects where the distance to (2, 6, 5) is greater than 3? \\
   & direction    & what are the names of the objects to my right?  \\
   & \textit{closest object}        & what is the name of the object that is closest to the cow?        \\
   & \textit{max direction} & what is the name of the object that is the most to my right?  \\
    
   & \textit{distance between}      & how far is the horse from you?\\
   & \textit{distance from position}      & what is the location 3 steps to your right?\\
   \bottomrule
\end{tabular}
}
\caption{
\vspace{-2pt}
Query clause types. We categorize the queries we can ask the agent into three separate classes. Within each class, there are several clause types. 
\textit{Italicized} clauses cannot be combined with others.}
\vspace{-3pt}
\label{tab:query_types}
\end{table*}

The embodied agent environment allows for a rich set of possible queries to ask the agent. We structure the types of queries we use into three main categories, as covered in Table~\ref{tab:query_types}. \textit{Property} queries are those which operate on the current state of the memory such as the current properties or locations of an object, and are given with an explicit relation.  Property queries with a single clause can be read directly from the database or text dump without any ``reasoning''. \textit{Temporal} queries are those which operate over spans of the memory such as the movement of object. \textit{Geometric} queries are those concerned with the geometric structure of the environment such as how far away two object are from each other.  Note that many queries are mixes of these types, and the categorization is blurred.

Within each query class, there are several different ``clause'' types. These can be combined as applicable into a multi-clause query, where the clauses are combined by an ``and'' or ``or'' conjunction randomly.  In addition, each clause can be negated by prepending the word ``not''. For example, ``what are the name of the objects that do not have the property brown and where the x coordinate is less than 4''.

The query, $Q$, is composed of one or both of the following representations: 
\textit{query\_text} ($Q_{t}$): a text representation of the query (e.g. ``find the reference\_objects with color brown''), \textit{query\_tree\_logical\_form} ($Q_{lf}$) : a tree logical form representation of the query (e.g. two clauses in an ``and'' or ``or'' query are connected by a parent in the tree).

Given the context and query, the agent should return some answer, $A$. Depending on the query type, we randomly ask for one of the following answer types in the query: \textit{name} (``what is the name of...''), \textit{properties} (``what are the properties of...''), \textit{location} (``what is the location of...''), \textit{distance} (``how far is...''), and \textit{count} (``how many...''). In general, we are interested in predicting the text answer such as ``brown'' for the query ``what is the color of the horse?''. However, we may also be interested in pointing to the relevant context objects or properties (in the structural context representation). Thus, for each query, we provide both the text answer as well as the relevant context node IDs (from $C_{R}$ and $C_{T})$.

\subsection{Data} With this data generation framework, we can create arbitrary amounts of simulated data. Each data sample contains a ($C$, $Q$, $A$) triple. There are several parameters of the world that affect the difficulty of question answering, including the size of the 3d gridworld (e.g., 15x15x15), the set of possible objects in the world (e.g., bob, alice, ...), and the set of possible properties that can belong to the objects (e.g. cow, horse, blue, green, ...). The number of time-steps and number of snapshots also crucially affect the difficulty of the problem. Similarly, the distributions over the queries (e.g., how many and what kind of clauses) can make the problem more difficult.

\section{Related Work}

\begin{figure*}[ht!]
\centering
\renewcommand{\arraystretch}{0}
\begin{tabular}{ccc}
\hspace{-3mm}
    \includegraphics[width=0.32\textwidth]{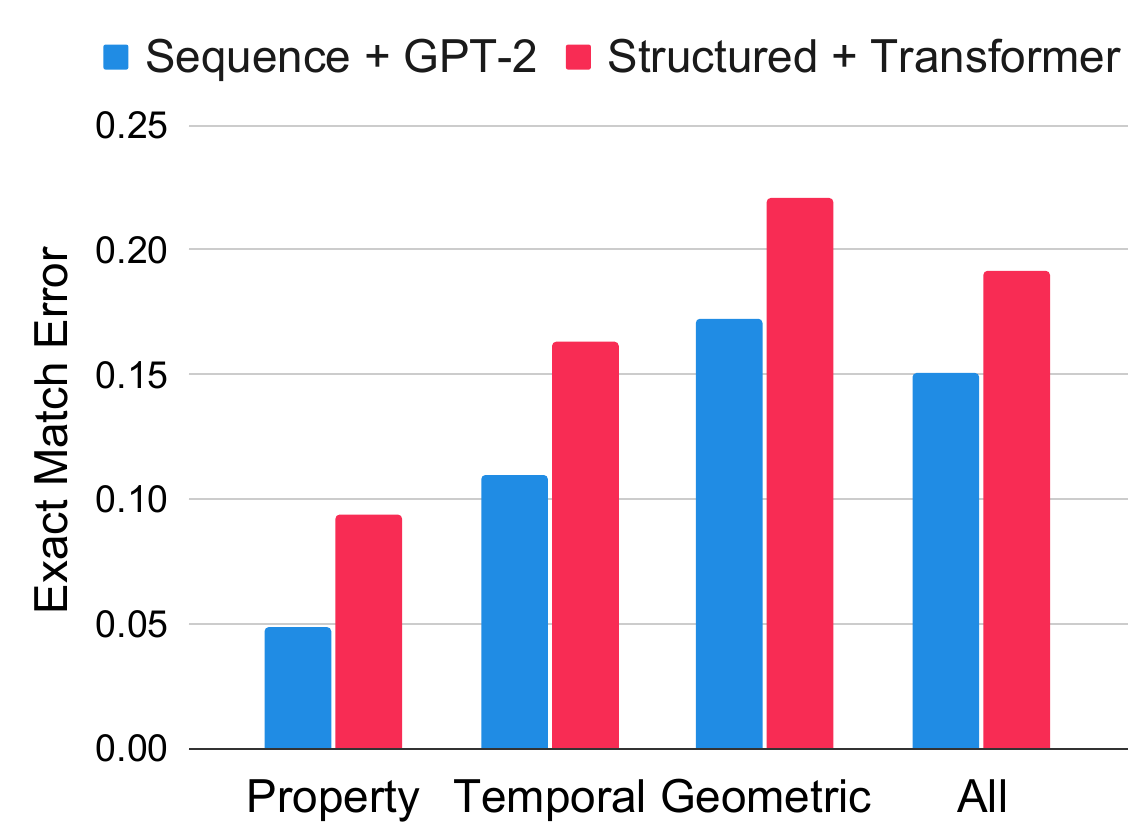}

&
\hspace{-4mm}

\resizebox{0.32\textwidth}{!}{
\includegraphics[]{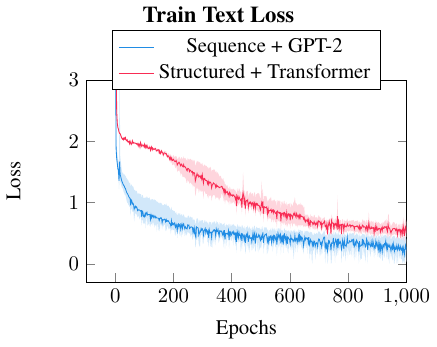}
}

\hspace{-6mm}

\resizebox{0.32\textwidth}{!}{
\includegraphics[]{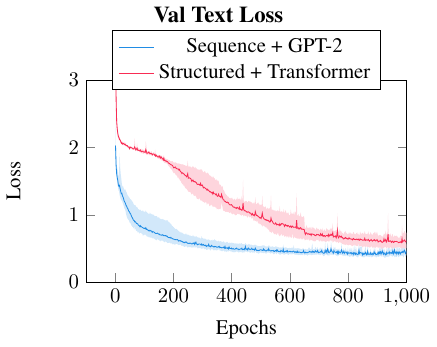}
}

\end{tabular}
    \caption{
    (left): Exact match error for the four different generated datasets. Sequence Context + GPT-2 outperform the Structured + Transformer method in all datasets.
    (middle, right): Loss curves for the All queries dataset.  We show the mean loss with min/max error bars over all hyperparameters for the first 1,000 epochs.   The pre-trained GPT-2 model learns much faster than the from-scratch relational model.
    }
    \label{fig:results}
    \vspace{-3pt}
\end{figure*}

Real-world QA datasets have long been used to test different aspects of ML model performance such as reading comprehension~\cite{Rajpurkar2016SQuAD1Q,Hill2016TheGP}, common-sense reasoning~\cite{Talmor2019CommonsenseQAAQ}, multi-hop reasoning~\cite{Yang2018HotpotQAAD}, and visual understanding~\cite{Agrawal2015VQAVQ,Hudson2019GQAAN}. 
While real-world datasets can provide reliable performance benchmarks and better approximate the problems faced by practitioners, synthetic datasets can allow for more control and the ability to isolate the exact limitations of current models.
Notably, bAbI~\cite{Weston2016TowardsAQ} is a set of toy QA tasks testing various reasoning abilities over short text stories that showed the limitations recurrent neural networks.
Since proposed, all bAbI tasks have been solved by novel memory architectures~\cite{Henaff2017TrackingTW,Dehghani2019UniversalT}.  An gridworld environment for embodied agents with language instructions for tasks is described in \cite{chevalier2018babyai}; our work here is complementary, giving question-answer pairs based on abstracted environment histories.

CLEVR~\cite{Johnson2017CLEVRAD} is a popular synthetic data for testing visual reasoning given text queries.
\cite{Yi2020CLEVRERCE} extends CLEVR to reasoning over temporal events in videos. %
Embodied question answering (EmbodiedQA) \cite{das2018embodied} proposes a task where an agent must navigate an environment in order to answer a question. VideoNavQA \cite{cangea2019videonavqa} was proposed in the EmbodiedQA domain to evaluate short video-question pairs.    Our work has elements of each these.  The agent is embodied, and might need to answer questions about its actions or hypotheticals, but does not need to act or change the current state of  the environment to answer (as in EmbodiedQA). In comparison to the EmbodiedQA dataset where the agent has to query the environment to get more information, our setting doesn't require the agent to interact and get more information.  As in  \cite{Yi2020CLEVRERCE}, the agent needs to be able to reason over spatio-temporal events, in our case, including its own actions.  As in CLEVR, we use programmatically generated queries to probe various reasoning modalities.  One large difference between this work and those is that we do not focus on computer vision.  While it is possible to render the scenes from our data generator, our goal is to be agnostic about perceptual modality and abstract away perceptual modeling, and to the extent possible, focus on the reasoning aspects of the data.  Within the vision community, other works have approached the VQA problem from this angle \cite{yi2018neural}.  %
\newline\indent Because the agent's abstracted world representation has a database-like structure, our work falls into the literature on ML for question answering on structured data, for example \cite{Pasupat2015CompositionalSP}. Our structured Transformer baseline is inspired by the literature on neural database representation, for example \cite{wang2019rat, yin2020tabert}, and references therein. Our LM baseline is inspired by the many works that flatten or otherwise textify databases, and use pretrained language models as bases for neural query executors, e.g. \cite{thorne2020neural, thorne2021database, liu2021tapex}.  There are other fundamentally different approaches than these for neural query execution, for example \cite{ren2020query2box}; our hope is that our data source is useful for exploring these. \cite{tuan2022towards} introduce a Transformer to generate responses to questions by reasoning over differentiable knowledge graphs in both task-oriented and domain specific chit-chat dialogues. Our structured neural memory baseline follows works such as \cite{locatello2020object,santoro2018relational}.    In this work, the relational ``objects' do not need to be discovered by the learning algorithm, and their properties explicitly given to the model to use in featurizing the objects.
Our work is most related to the pigpen environment of \cite{zellers2021piglet}.  In comparison to that work, ours uses more impoverished templated language, but has a larger and more flexible space of queries.  The orientation of our work is also different: in \cite{zellers2021piglet} the learner is given only a few labeled QA examples (or dynamics prediction examples) but can use many ``unsupervised'' state-transitions to build a world-dynamics model.  In our work, we allow large numbers of labeled context-query-answer examples; but the difficulty of the queries makes the task non-trivial. 

TextWorld \cite{cote2018textworld} and and QAit \cite{yuan2019interactive} are text-based environments for game play and question answering that require interactive reasoning. The main difference is that our generator is grounded in a 3D gridworld scene, and there is both a user and interactive agent.

We build our data generator on top of the Droidlet agent \cite{pratik2021droidlet}, in part using the grammar in \cite{srinet2020craftassist} to generate the queries and using the Droidlet agent memory to execute them.  This work points the way towards using neural networks to execute the functions of the Droidlet agent memory (and hopefully can be used as a resource for training other agent-memory models).

\label{sec:experiments}
\begin{figure*}[t]
    \centering
    \includegraphics[width=0.85\textwidth]{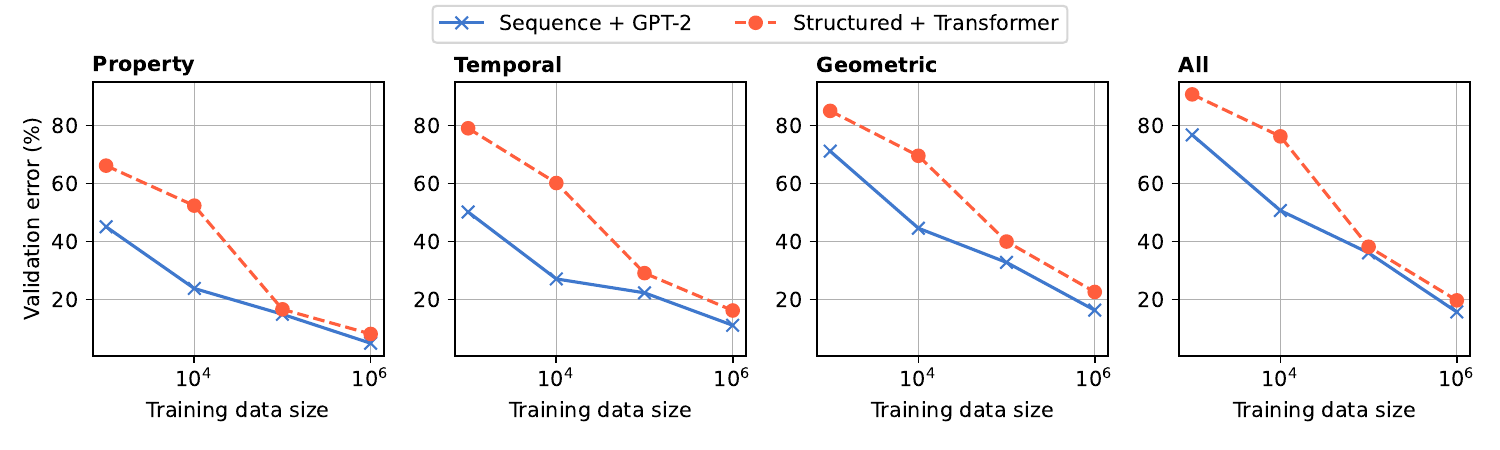}
    \vspace{-3pt}
    \caption{
    Validation set exact match error using varying amounts of training data. The pre-trained GPT-2 model is particularly useful when the number of training samples is small, and when reasoning over proprty queries. On the other hand, non-pre-trained GPT-2 style models do not learn, see Table \ref{tab:gpt_variations}. Both models require more than 100K training samples to achieve the best results. %
    }
    \label{fig:results2}
    \vspace{-5pt}
\end{figure*}
\vspace{-7pt}
\section{Experiments}
\vspace{-7pt}
Since the agent's memory or state can take different forms (sequence and structured), we compare two separate models for answering queries about the context of the world. We consider four different datasets for our experiments, as covered in Table~\ref{tab:query_types}: Property queries, Temporal queries, Geometric queries, and All queries, where \textit{All queries} is the three previous categories combined (each query type has roughly the same likelihood of occurring - we provide the configuration files in the code). 

Each of these except for Properties (which do not require temporal information) is generated using two time snapshots with 50 world steps, which gives enough steps for actions to occur. Properties queries use one snapshot and zero world steps.  For all queries, we place five NPCs in the world, one of which is a ``player'' that might have given a command.  For all query types, we choose the world to be 15x15x15.  

\subsection{Models}
\label{sec:methods}
We generate data as described in the Environment, Queries, and Data section, and analyze the performance of some baseline models trained on this data.
\paragraph{Text Sequence Context.}  Since the text sequence form of the context is English text, we use a language model to read the context ($C_{t}$) and query ($Q_{t}$), and predict the correct answer tokens sequentially (if there are multiple outputs, they are ordered alphabetically). We use the pretrained GPT-2 small model \cite{radford2019language} from the HuggingFace library \cite{wolf2019huggingface} (licensed under the Apache License 2.0) 
to predict all relevant tokens sequentially:
\begin{align}
    \bm{\hat{W}}&=\textrm{GPT2}([C_{t},Q_{t}]),
\end{align}
where $\bm{\hat{W}} \in \mathbb{R}^{L \times V}$ is a soft-max normalized matrix of sequence length $L$ and vocabulary size $V$, and $[\,]$ is the concatenation operation. 
This model is fine-tuned using a sum of the cross entropy between each token prediction $\bm{\hat{W}_i}$ and ground truth token $\bm{W_i}$:
\begin{equation}
    \mathcal{L}_{text} = -\sum_{i \in S} \sum_{j \in V} \bm{W}_{ij} \log \bm{\hat{W}}_{ij}.
    \label{eq:text_loss}
\end{equation}
\paragraph{Structured Context.} While the text Sequence Context is useful in that it allows easy application of standard pre-trained models, it may be that for certain tasks other representations are more appropriate.  We also show results with simple models that are designed around the relational structure of the context. Given a set of $\rho$ reference objects and $\tau$ triples, we first featurize the nodes using a learned convolutional layer given the word, float, and hash values as outlined in the Environment, Queries, and Data section. The output of the featurizer layer gives reference object embeddings $C_{R} \in \mathbb{R}^{\rho \times d}$, and triple embeddings $C_{T} \in \mathbb{R}^{\tau \times d}$. Similarly, text queries $Q_{t} \in \mathbb{R}^{r \times d}$ are created using a learned lookup table from the query tokens. We use a Transformer \cite{vaswani2017attention} encoder to process the context and query. The output of the encoder is then used to predict both: the text answer as well as the relevant context memory IDs. We use a Transformer decoder for the text prediction, and a simple linear layer to predict the memory values:
\vspace{-5pt}
\begin{gather}
    (C_{T}',C_{R}',Q'_{t}) = \textrm{Encoder}([C_{T},C_{R},Q_{t}])\\
    \bm{\hat{m}}= \textrm{MemidDecoder}([C_{R}',C_{T}'])\\
    \bm{\hat{w}}= \textrm{TextDecoder}([C_{T}',C_{R}',Q'_{t}]),
\end{gather}
where $\bm{\hat{m}} \in \mathbb{R}^{\rho}$ is predicted relevant memids, and $\bm{\hat{w}} \in \mathbb{R}^{V}$ represents predicted relevant text tokens. This model is trained using a cross-entropy text token loss (eq \ref{eq:text_loss}) and cross-entropy memid loss from the relevant memids $\bm{m} \in \mathbb{R}^{\rho}$:
\vspace{-5pt}
\begin{equation}
    \mathcal{L}_{memid} = -\sum_{i=1}^{C} \hat{m}_i \log \left({m_i}\right).
\end{equation}
\vspace{3pt}
The total loss is a weighted sum of the two losses: $\mathcal{L}_{text}$ + $\lambda \cdot \mathcal{L}_{memid}$, where $\lambda=0.5$.

In the Sequence Context, the entities are represented as a set, with defined relationships between them. Therefore, encoding temporal information is not straightforward. To do so, we add a special ``time'' embedding to each Sequence Context node. That is, for timestep 0, we add the $time$=0 embedding, for timestep 1, we add the $time$=1 embedding.

For the Sequence Context, the model will return the relevant memory IDs, a text sequence answer, or both. The text output prediction is shown in Fig.~\ref{fig:structured_model}.
\paragraph{Relational Embedding.} One important form of structure in data is relations where entities are connected to other entities. In our data, for example, triple nodes are connected to object nodes by R\_id. Since the vanilla Transformer without positional embeddings treats input tokens as a set, it lacks a mechanism for taking account of this relational information. Thus, we propose a novel way of encoding relational information directly into the self-attention mechanism of the Sequence Context Transformer model. Specifically, we add an extra term to the softmax attention:
\vspace{2pt}
\begin{equation}
    a_{ij} = \text{Softmax}(q_i^T k_j + q_i^Tr_{ij})
\end{equation}
\vspace{2pt}
Here $r_{ij}$ is a \emph{relation embedding} vector corresponding to the relation type between token $i$ and $j$. Relation embeddings can be used to encode various types of relations. We note that the commonly used relative position embeddings~\cite{sukhbaatar2015end,shaw2018self} are a special case of relation embeddings where $r_{ij}=e_{i-j}$; more sophisticated relation embeddings have appeared in \cite{bergen2021systematic}.   

\subsection{Model and Training Details}
All our models are trained using Adam \cite{kingma2014adam} for 5,000 epochs, where each epoch is over a chunk of 10,000 training samples. Since we are generating the data, we vary the training samples from 1k to 1M, and use a validation set of 10k samples. We use a linear warmup of 10,000 steps and cosine decay \cite{LoshchilovH16a}. For the GPT2 model, we consider learning rates \{1e-4, 5e-4, 1e-5\} using a batch size of 32. For the structured model, we consider learning rates \{1e-4, 5e-4, 1e-5\}, batch size 32, layers \{2, 3\}, and embedding dimensions \{256, 512\}. Hyperparameters were chosen with arbitrary initial values and increased until validation performance decreased or resources were depleted. The best performing structured model has 74.58M parameters, whereas GPT-2 small has 325.89M. 
All words are encoded with the GPT-2 tokenizer.

\subsection{Results}
Fig.~\ref{fig:results} (left) shows the results for the four different dataset versions we consider. We report the exact match error for all data splits. That is,
\vspace{-5pt}
\begin{equation}
    \textrm{Exact Match Error} = \frac{1}{n} \sum_{i=1}^{n} I\left(\bm{y}^{i}\neq\bm{\hat{y}}^{i}\right),
\end{equation}
where $\bm{y}^{i} \in \mathbb{R}^N$ are the $N$ ground truth tokens for sample $i$, and $\bm{\hat{y}}^{i}$ are the top $N$ predicted tokens.  Fig.~\ref{fig:results} (right) shows the loss curves for the All queries dataset.

For property and geometric queries, the Sequence Context + GPT-2 method performs the best. Since GPT-2 is pre-trained on millions of samples, it can easily learn basic property queries such as ``what is the color of bob?''. GPT-2 has a near zero test set loss for the properties queries. Geometric queries, while difficult for both models to answer is solved more effectively by GPT-2.

\begin{figure}[ht]
    \centering
    \includegraphics[width=0.92\columnwidth]{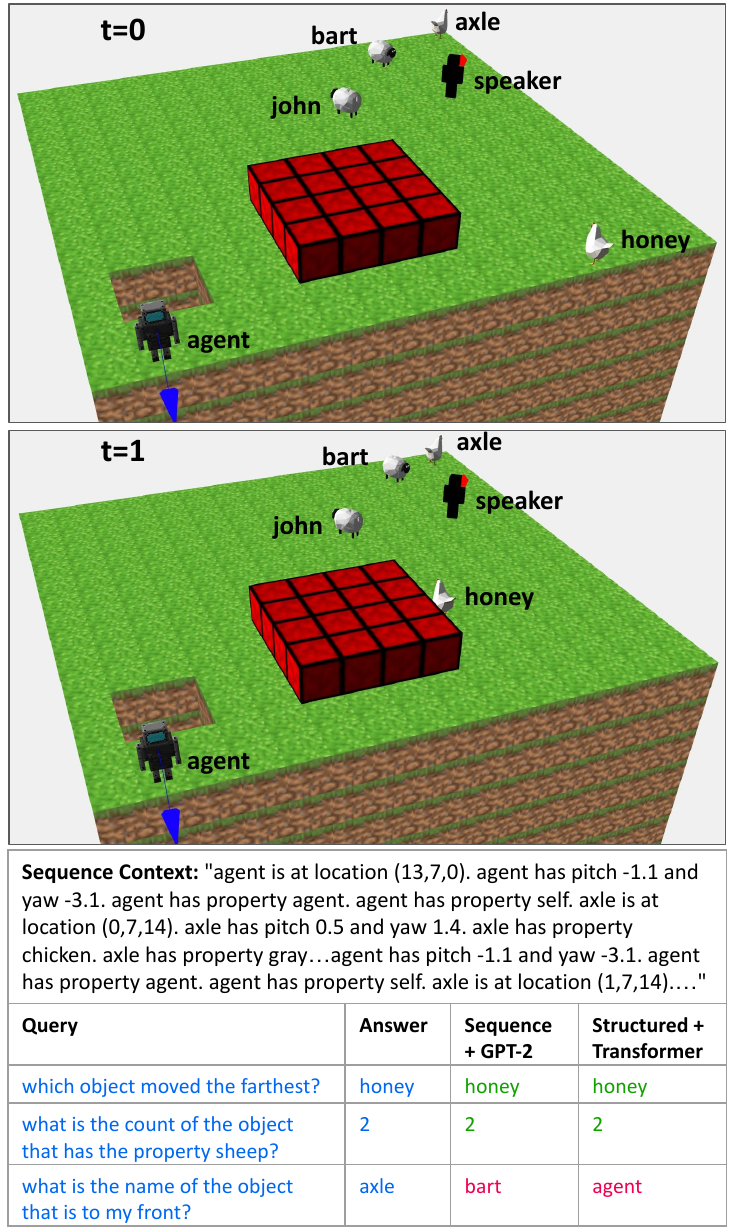}
    \vspace{-5pt}
    \caption{
    Sample scene with two snapshots. The chicken named ``honey'' moves from the bottom right toward the center, and some other objects move slightly. Both models correctly answer the query ``which object moved the farthest?''.}
    \label{fig:examples}
    \vspace{-5pt}
\end{figure}

Table~\ref{tab:gpt_variations}  (top) shows GPT-2 model variation studies for the All queries dataset. Notably, we test the performance of a Sequence Context + randomly initialized GPT-2 model (i.e. one with the same architecture, but not pretrained). We see that the performance is worse than that of the Structured Context + Transformer. This indicates that the primary reason for the GPT-2 model achieving the best results is due to the pretraining. Table~\ref{tab:gpt_variations} (bottom) shows a Structured Context + Transformer method variation for the All queries dataset. We consider an alternative to the relational embeddings proposed in 
the Models section, 
which is adding random hash embeddings to the nodes, or tokens, which are connected to each other in the Sequence Context representation. This method performs slightly worse than relational embeddings, hinting that a more explicit representation of the context structure is important. Fig.~\ref{fig:examples} shows a 2-snapshot scene from our test set.

Finally, Fig.~\ref{fig:results2} shows the result of our baseline models when using varying amounts of training samples. Since our data source can generate arbitrary amounts of data, it's important to understand how much data is needed to solve certain problems. Notably, most of the datasets require at least 100,000 samples to achieve a reasonable validation loss. In addition, the Sequence + GPT-2 model significantly outperforms the Structured + Transformer model on the datasets with small amounts of training samples.

\begin{table}[]
\fontsize{9}{9}
\setlength{\tabcolsep}{4pt}
\centering
\resizebox{\columnwidth}{!}{
\begin{tabular}{llccc}
\toprule
Model   & Variation                  & \multicolumn{1}{l}{Train Loss} &  \multicolumn{1}{l}{Test Loss} & \multicolumn{1}{l}{Test Err} \\ \midrule
\multirow{3}{*}{GPT-2} & sm + rand init & 0.361                                                          & 1.912                               & 47.1\%                         \\ 
 & sm + pretrain  & 0.015                                                       & 0.710                             & 14.9\%                         \\ 
 & med + pretrain & 0.012                                                       & 0.635                             & 13.8\%                         \\ \midrule
\multirow{2}{*}{Transfomer} & relational emb & 0.230                                                         & 0.921                             & 19.7\%                         \\ 
 & random hash           & 0.630                                                        & 1.393                            & 30.0\%                       \\ \bottomrule
\end{tabular}
}
\vspace{-5pt}
\caption{
Variations of models (sm=small, med=medium). Pre-training is a key component of the GPT-2 model success, even when given large numbers of training examples.
 Relational embeddings result in a slightly lower test loss than random-hash in the graph-structured models.
}
\vspace{-7pt}
\label{tab:gpt_variations}
\end{table}

\section{Conclusion}

In this work, we introduced a framework for generating world state contexts with agents, queries, and their answers. This provides researchers with a flexible sandbox for training and testing reasoning in embodied agents. Notably, our sequential data format lets us easily evaluate the ability of large language models such as GPT to understand a physical world.

We show baseline results with two representations of the world state context: a pure text sequence representation that can be used by any off-the-shelf language model, as well as a structured representation of the entities in the world.
We demonstrate the ability of the models to answer queries in several query domains, such as temporal or geometric.

We emphasize that our goal in this work is not to create a fixed static data set, but a resource for generating data.  Many of our experimental choices (w.r.t. environment and query difficulty) were aimed to allow the baseline models some traction, but not able to solve the tasks completely.  

In particular, we find it likely that large pre-trained language models could do better than the GPT-2 base models we used with the described data generation parameters.  On the other hand, the difficulty of the problem can be scaled trivially by increasing the number of possible reference objects and properties, by increasing the number of reference objects in the world and increasing the number of instantiated properties; by making the world bigger, and by increasing the time-length of episodes and the number of snapshots recorded.  If nothing else, these changes would quickly lead to the context being too large to fit in the memory of a standard LM, and necessitating a large-memory LM \cite{lample2019large, rae2019compressive, beltagy2020longformer}; or leading to other approaches e.g. \cite{ji2017dynamic}.  Other, more subtle difficulties can be introduced in straightforward ways by reducing full observability, or otherwise asking queries that cannot be answered with the information in the context, requiring agents some amount of meta-cognition and/or environment actions to find the answers, as in \cite{das2018embodied}.

In future work, we plan to introduce more query types, including arithmetic and hypotheticals.   We hope that researchers will use the data generator as a flexible resource for augmenting their own agent training or LM pre-training, or for exploring new models for database reasoning.

 \clearpage 
 \bibliography{neural_memory}

\clearpage 
\section{Appendix}

\subsection{Scene, Query, and Training Details}
\begin{figure*}[]
    \centering
    \includegraphics[width=0.5\textwidth]{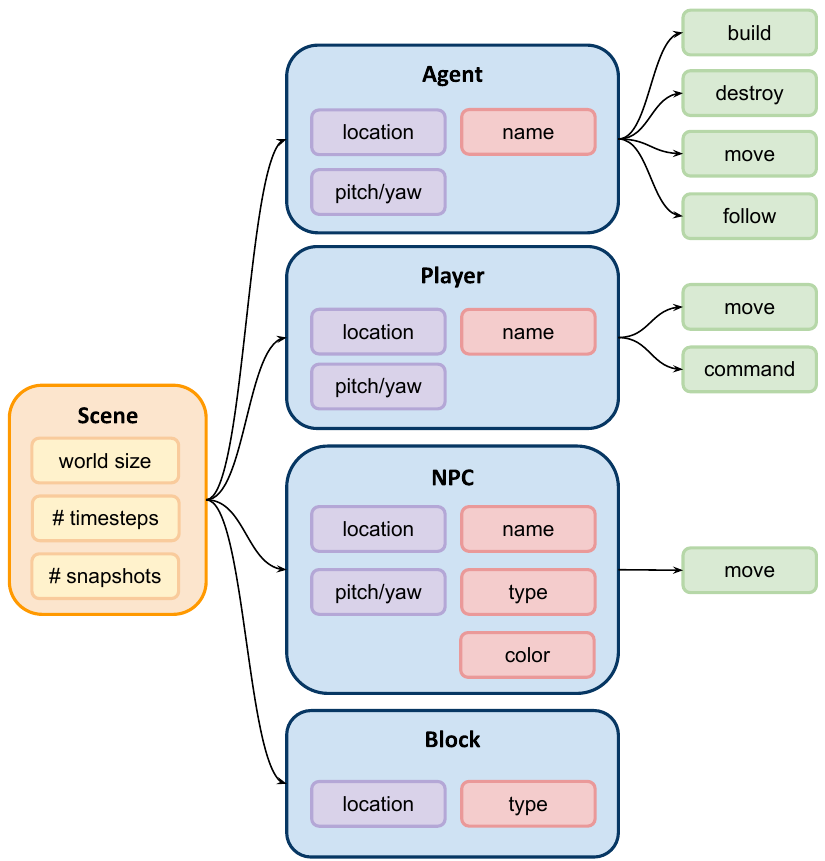}
    \caption{Scene context ($C$) components. Our data source allows us to create highly modular and variable scenes. We first select a gridworld size, number of timesteps and number of snapshots, and then randomly generate a number of objects in the world, including the agent, the player, NPCs, and blocks. Certain objects can perform actions such as move and build.} 
    \label{fig:build_scene}
\end{figure*}

\textbf{Scene generation:} Figure~\ref{fig:build_scene} shows the components of building a scene in our 3d gridworld. Parameters of the scene generator include: world size, number of timesteps, number of agents, number of players, number of NPCs, (names, properties, locations, pitch/yaw of the NPCs and agent), and actions. In our experiments, we use 1 agent, 1 player, 4 NPCs generated from a set of 254 names (Abigail, Ace, Adam, ...), 5 NPC types (cow, pig, rabbit, chicken, sheep), 6 NPC colors (brown, white, black, mottled, pink, yellow), and up to 5 shape types for block objects (hole, cube, hollow\_cube, rectanguloid, hollow\_rectanguloid, sphere, spherical\_shell, pyramid, square, rectangle, circle, disk, triangle, dome, arch, ellipsoid, hollow\_trianlge, hollow\_rectangle, rectanguloid\_frame). At each timestep, the agent, player, and NPCs can perform actions (move, build, destroy, follow, command). Some of these actions create new blocks or destroy existing ones. Note that the actions are not explicitly recorded in the context $C$. Actions such as ``move'' are implictly recorded by the object changing its location, building by the existence of new block objects, etc.

\begin{figure*}[]
    \centering
    \includegraphics[width=0.74\textwidth]{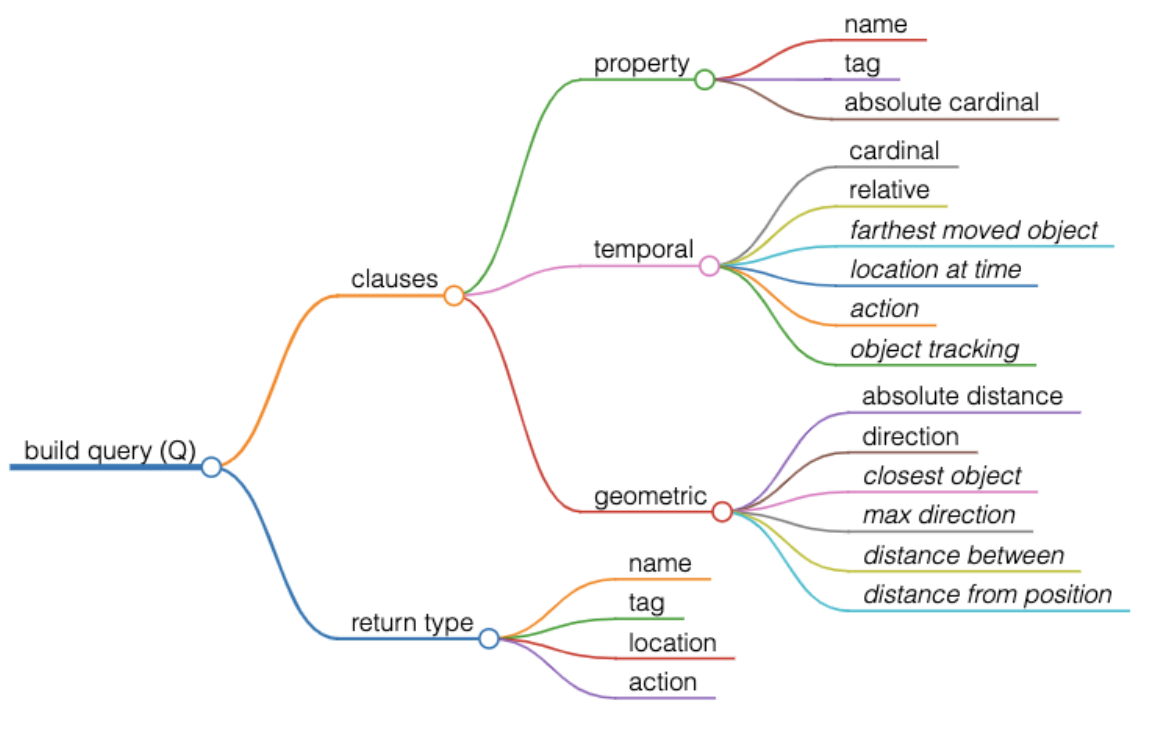}
    \caption{Query Generation Flowchart. \textit{italicized} clauses in the tree are standalone clauses, and cannot be combined with others. Non-italicized clauses can be combined with conjunctions to form multi-clause queries. Some return types are not compatible with certain clauses.}
    \label{fig:query_gen}
\end{figure*}

\textbf{Query specifications:} %
Figure~\ref{fig:query_gen} shows an overview of the query generation process. Currently there are 13 different clause types, some of which can be combined with others to create multi-clause queries. There are 4 different standard return types for the query: ``name'' (e.g. ``what is the name of ...''), ``tag'' (e.g. ``what is a tag of ...''), ``count'' (``how many ... are there ''), and ``location'' (`` what is the location of ...''). Not every one of these is applicable for each of the 13 clause or query types. Specifically:
\vspace{-2pt}
\begin{itemize}
    \item For agent action queries (what did you do?), the return is always an action name, and none of the above entity-centric questions.
    \item For geometric minimax queries (closest object, max direction), the return is always the name or tag of the object.
    \item For all other query types where only one output is possible (based on the query alone, and not on the state or history of the world), count queries are not allowed. 
\end{itemize}

\textbf{Compute, resources, and availability}
Each model is trained on an internal cluster, using 4-16 Nvidia V100 GPUs. Training takes roughly 72 hours for each model. The best performing model has 74.58M, whereas the GPT-2 small model has 325.89M parameters. All hyperparameter configurations and PyTorch \cite{paszke1912pytorch} code will be released to both generate the data and train the models.

\subsection{Detailed analysis}
\label{sec:detailed}
\begin{figure*}[]
\renewcommand{\arraystretch}{0}
\begin{tabular}{ccc}

\resizebox{0.45\textwidth}{!}{
\includegraphics[]{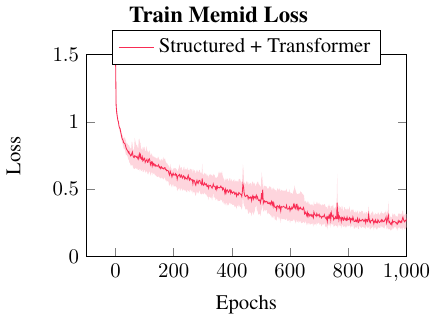}
}

\hspace{-6mm}

\resizebox{0.45\textwidth}{!}{
\includegraphics[]{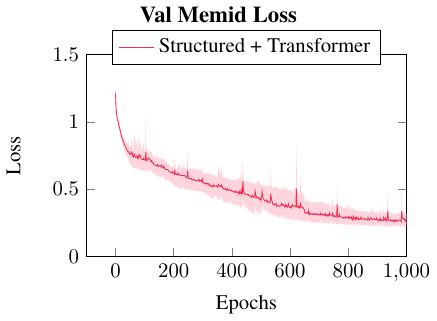}
}

\end{tabular}

    \caption{Memid loss for the All queries dataset. This loss is minimized jointly with the text loss for the Structured+Transformer model. The memid prediction task is a simpler task than predicting the text tokens since the model only has to select from objects present in the scene vs all possible tokens. 
    }
    \vspace{5pt}
    \label{fig:memid_loss}
\end{figure*}

\textbf{Memid loss.} While the main objective is to predict the text answer, for the structured context representation, we are also interested in predicting relevant context node IDs (i.e. ``memids''). Figure~\ref{fig:memid_loss} shows the memid loss ($\mathcal{L}_{memid}$) for the All queries dataset. Note that the GPT-2 model does not operate on the structured memory nodes, so there is no memid loss. This loss is lower than the text loss since the model only has to output the correct IDs from the small set of objects in the world, whereas the model has to predict the correct text tokens from all possible tokens (50,257 for the GPT-2 tokenizer). We observe that without the memid loss, the text prediction performance suffers.

\begin{figure*}[t]
    \centering
    \setlength{\fboxsep}{0pt}%
    \setlength{\fboxrule}{1pt}%
    \includegraphics[width=0.8\textwidth]{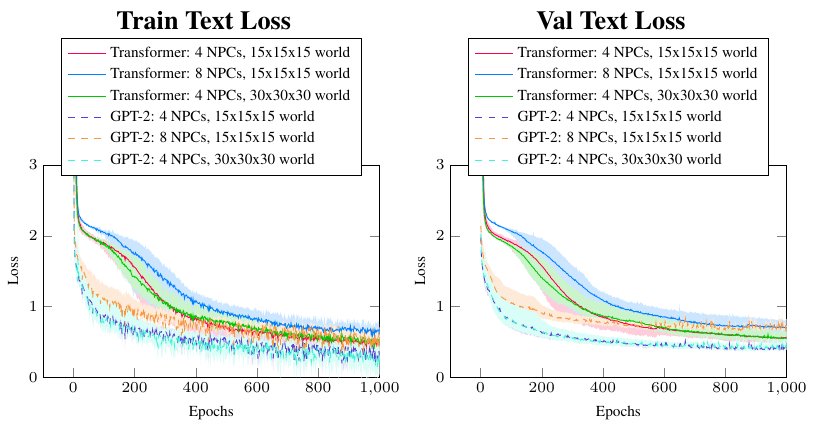}
    \vspace{-5pt}
    \caption{Varying dataset complexity. We vary two dataset parameters to make the queries more difficult: increasing the number of NPCs from 4 to 8, and increasing the world size from 15x15x15 to 30x30x30. We observe that as the number of NPCs grows, the queries become more difficult to answer, but the Structured+Transformer model begins to outperform the Sequence+GPT-2 model (on this dataset, the best GPT-2 val loss is 1.055, and the best Transformer val loss is 0.915).}
    \label{fig:dataset_complexity}
\end{figure*}

\textbf{Generating more difficult datasets.} In the main manuscript, we show the results of baseline models where every scene has 4 NPCs in a 15x15x15 gridworld. We can generate scenes according to our specifications, making it simple to test the model capabilities on samples of varying difficulty. Figure~\ref{fig:dataset_complexity} shows the results when adjusting two parameters of the world for the All queries dataset: larger gridworld size, and more NPCs. The baseline is a 4 NPC, 15x15x15 world. The first variation is a 4 NPC, 30x30x30 world, and the second is a 8 NPC, 15x15x15 world. The performance remains roughly the same for the larger world size. We attribute this to the fact that the models already have trouble solving the geometric queries in the smaller world size, so increasing it doesn't degrade the performance. However, the performance does decrease when we increase the number of NPCs. Notably, the Structured+Transformer model begins to outperform the Sequence+GPT-2 model in this setting. This is likely in part because in a world with more NPCs, the sequence length becomes longer than 1024 tokens, which is the length GPT-2 is trained on. As we clip the context at 1024 tokens, it is not able to answer queries where the relevant information is outside of the 1024 tokens. This highlights one inherent issue with naively using a pre-trained language model on this task. This problem will also be exacerbated when we increase the number of timesteps.  However, note the literature on ``long-memory'' Transformer text models, e.g. \cite{lample2019large, rae2019compressive, choromanski2020rethinking, beltagy2020longformer, Sukhbaatar2021NotAM}. 

\begin{figure*}[]
    \centering
    \includegraphics[width=1.0\textwidth]{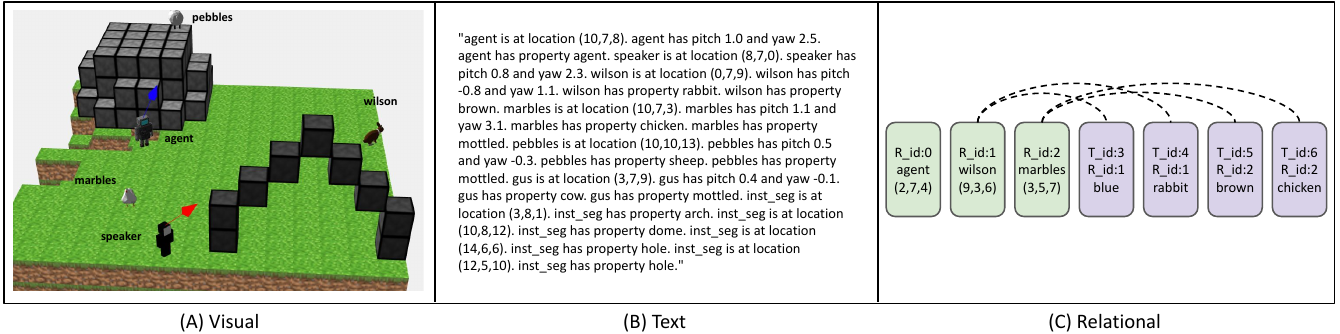}
    \caption{Three different data modalities available from our data generator for a single scene. (A) visual: not used in our models since we're primarily interested in learning memory representations. (B) text: flattened text used by the Sequence + GPT-2 model. (C) relational: structured object and property nodes used by the Structured + Transformer model.}
    \label{fig:data_representations}
\end{figure*}

There exist many similar datasets, with key differences in terms of the types of inputs available, and whether there is an active agent. Table~\ref{tab:dataset_comparisons} shows a comparison of our dataset vs. existing ones. The key differentiation is that we can represent the data in several forms: visual, relational, and text, as show in in Figure~\ref{fig:data_representations}.

\textbf{Out of domain generalization.} In real world scenes, new objects and properties may arise at any given time. For example, the agent may encounter an animal with a new color that we've never seen before, but it should still be able to answer queries about it. To understand how the model performs in this setting, we generate a new Properties dataset test set with all new color properties that were not seeing during training (e.g. green , orange). Table~\ref{tab:compositional} shows the results on the original colors Properties dataset vs the new colors Properties dataset. The Sequence+GPT-2 model generalizes significantly better than the Structured+Transformer model that was not pretrained.

\subsection{Minecraft Setting}
Minecraft is a suitable environment for agent reasoning for a variety of reasons. \cite{szlam2019build} argue that the constraints of the Minecraft world (e.g. 3-d voxel grid, simple physics) and the regularities in the head of the distribution of in-game tasks present a rich setting for natural language and world model understanding. Concurrent with our work, two works train video-based agents in the Minecraft setting. \cite{fan2022minedojo} introduce Minecraft simulation suite with thousands of open-ended tasks and an internet-scale knowledge base with Minecraft videos, tutorials, wiki pages, and forum discussions. They also introduce an agent that uses pre-trained video-language models as a reward function. \cite{baker2022video} train a video-model to predict the next action based on a scene in Minecraft.

\subsection{Additional Examples}

\begin{figure*}[]
    \centering
    \setlength{\fboxsep}{0pt}%
    \setlength{\fboxrule}{1pt}%
    \fbox{\includegraphics[width=0.85\textwidth]{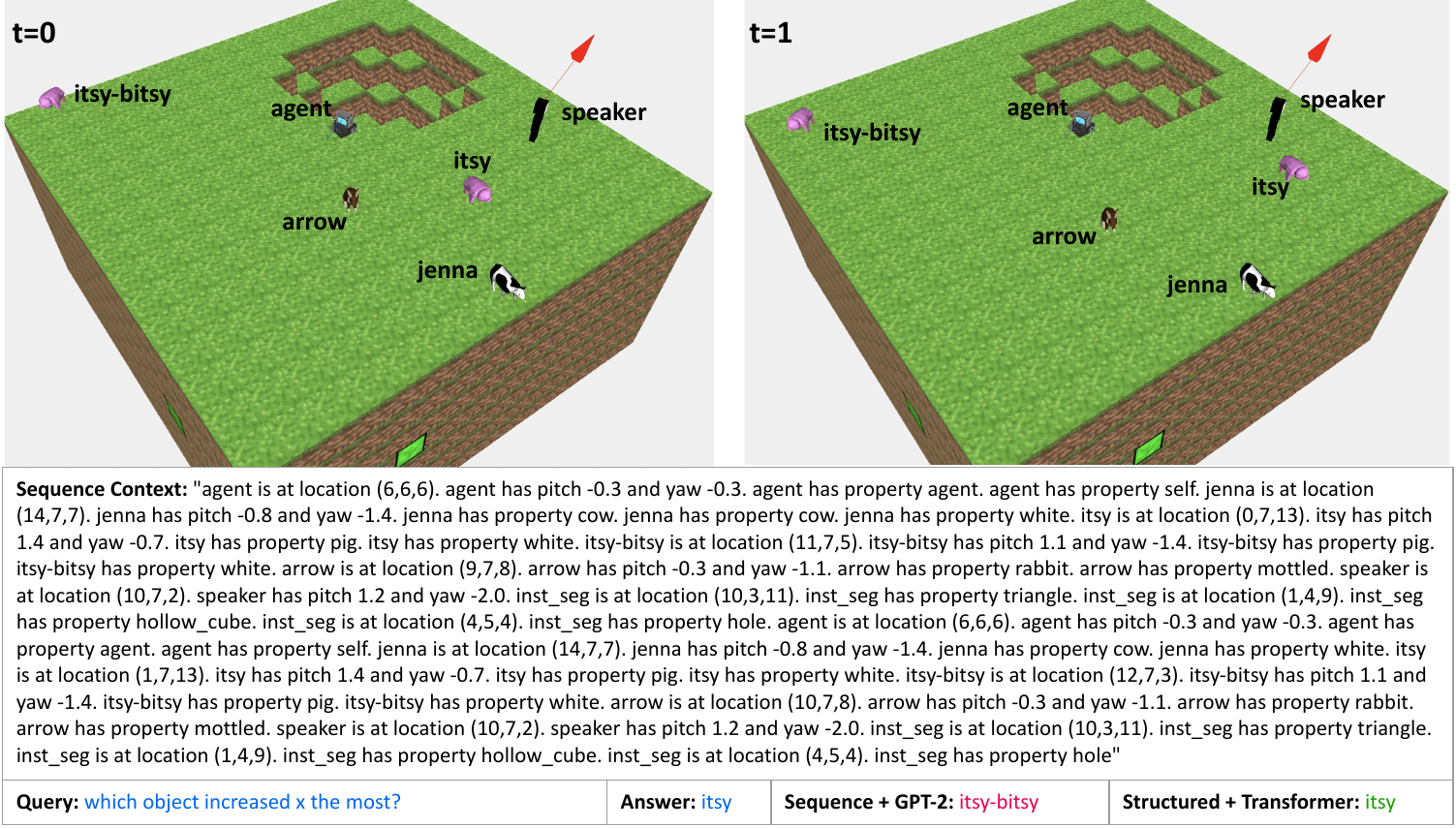}}%
    \,
    \fbox{\includegraphics[width=0.85\textwidth]{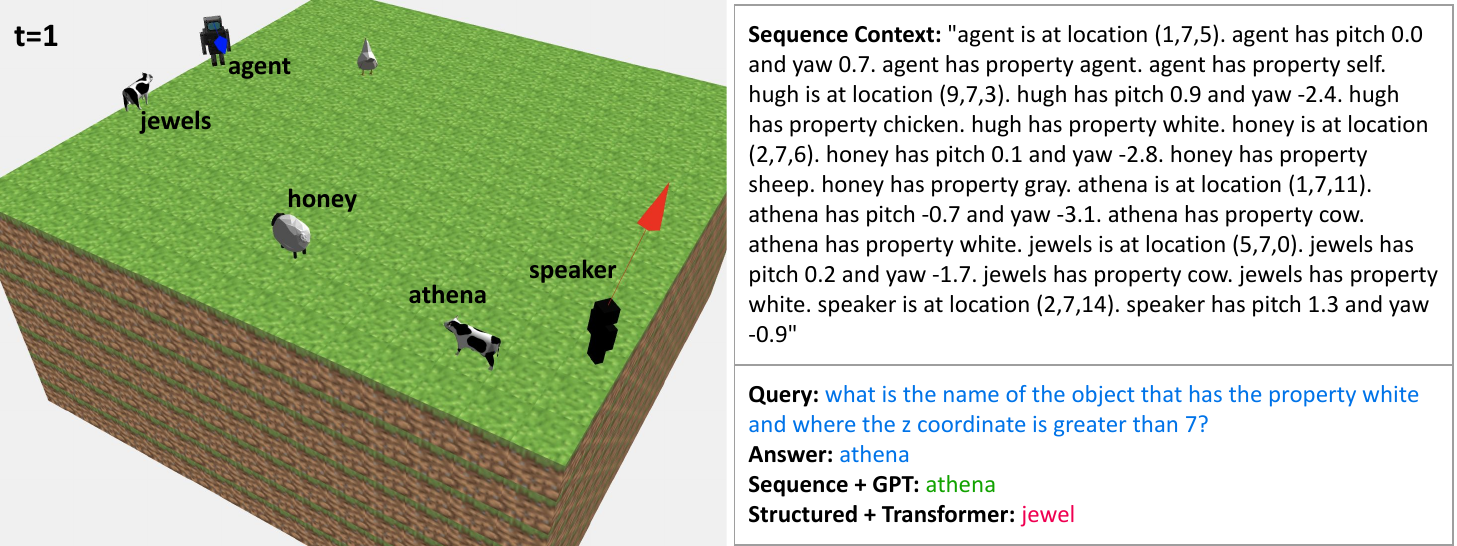}}%
    \,
    \fbox{\includegraphics[width=0.85\textwidth]{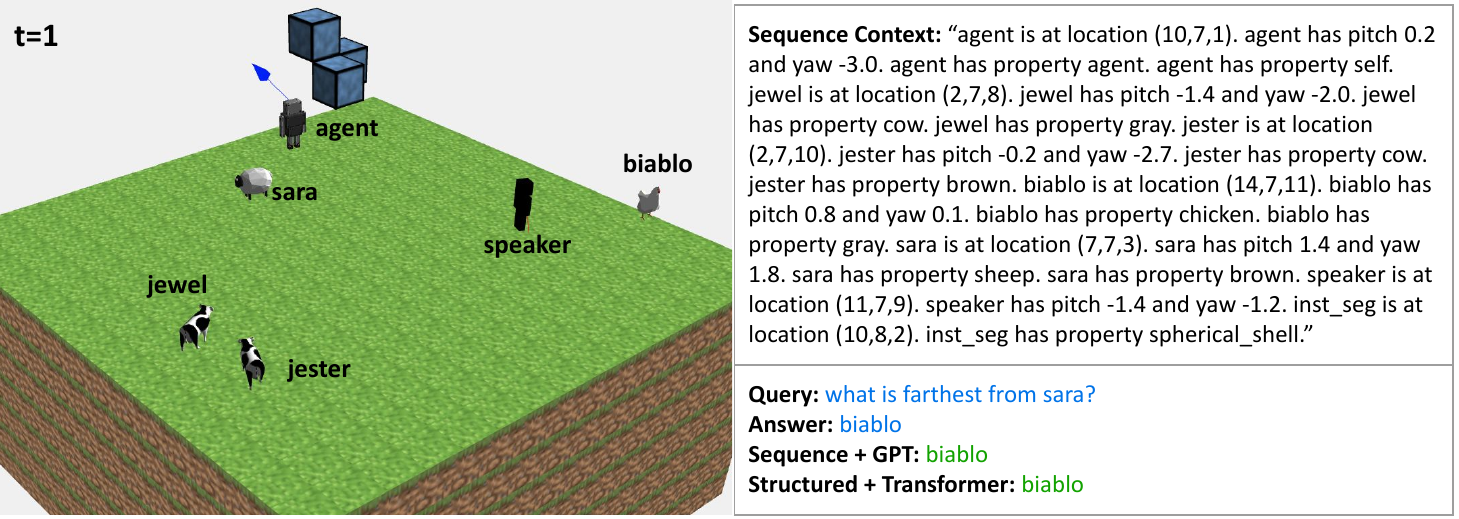}}%
    \caption{
    \textbf{Top}: Temporal query. In this example, both of the ``pig'' NPCs increased their x value, but the one named ``itsy'' increased the most. The GPT-2 model could not accurately predict the right answer.
    \textbf{Middle}: Multi-clause property query. In this example, but ``athena'' and ``jewel'' have the property ``white'', but only athena has a z coordinate greater than 7. The Transformer model could not accurately predict the right answer. Note that in queries such as this which don't require temporal reasoning, we consider the most recent timestep (t=1 in our expriments).
    \textbf{Bottom}: Geometric query. In this example, the NPC named ``biablo'' is the farthest from ``sara'' in the gridworld.
    }
    \label{fig:more_examples}
\end{figure*}

Figure~\ref{fig:more_examples} shows more examples of our environment with predictions from the two baseline models. Neither the Sequence+GPT-2 nor Structured+Transformer model are able to accurately answer many of the queries, motivating the need for more advanced modeling techniques. While these examples show basic gridworld examples, we can generate arbitrarily more difficult worlds, as described in Section~\ref{sec:detailed}. These examples demonstrate a small subset of possible worlds and queries generated from our data source. The potential scenes and queries that researchers can create are extensive, and can be carefully tailed toward the specific needs of a particular agent being trained. 

\clearpage

\begin{table*}[]
\centering
\begin{tabular}{l|c|c|cll}
Dataset    & Available data types & Embodied Agent    & Dataset size \\ \hline
CLEVR  \cite{Johnson2017CLEVRAD}    & Visual& & 1M             \\
EmbodiedQA \cite{das2018embodied} & Visual& \checkmark &   9K   \\
bAbI  \cite{Weston2016TowardsAQ}     & Text  & &      20K* \\
BabyAI  \cite{chevalier2018babyai}     & Text  &\checkmark & 5K*     \\
TextWorld \cite{cote2018textworld}  & Text    & &   150K \\
PIGPeN  \cite{zellers2021piglet}   & Text  &  &280k       \\
QAit \cite{yuan2019interactive}     & Text  &  &    500* \\ \hline
Ours       & Visual, Text, Relational  & \checkmark &   1M* 
\end{tabular}
\caption{Comparison of different QA datasets. ``*'' specifies that an unlimited amount of samples can potentially be generated.}
\label{tab:dataset_comparisons}
\end{table*}
\begin{table*}[t]
\centering
{\renewcommand{\arraystretch}{1.1}
\begin{tabular}{l|cc|cc}
\toprule
      & \multicolumn{2}{c|}{Original Colors}  & \multicolumn{2}{c}{New Colors}       \\ \cline{2-5} 
      & \multicolumn{1}{l|}{Loss}  & Exact Match Err & \multicolumn{1}{l|}{Loss}  & Exact Match Err \\ \hline
Sequence+GPT-2   & \multicolumn{1}{l|}{0.060} & 5.0\%  & \multicolumn{1}{l|}{0.647} & 16.3\%  \\ \hline
Structured+Transformer & \multicolumn{1}{l|}{0.092} & 9.4\%  & \multicolumn{1}{l|}{0.704} & 37.0 \%
\\ \bottomrule
\end{tabular}}
\caption{Out of domain generalization on the Properties dataset. The ``Original Colors'' test dataset uses the color properties such as ``white'' that the models were trained on. The ``New Colors'' test dataset exclusively uses new color properties such as ``green'' and ``purple'' that were never seen during training. The Structured+Transformer model sees a significant reduction in performance when introducing new color properties, while the Sequence+GPT model is relatively robust.}
\label{tab:compositional}
\end{table*}

\end{document}